\documentclass[]{llncs}
\usepackage{etoolbox} 
\makeatletter
\patchcmd{\ps@headings}{\rlap{\thepage}}{}{}{}
\patchcmd{\ps@headings}{\llap{\thepage}}{}{}{}
\makeatother
\usepackage{graphicx}
\usepackage{multirow}
\usepackage{hyperref}
\usepackage[frozencache=true,cachedir=minted-cache]{minted}
\UseRawInputEncoding
\begin{document}
\title{Sexism Prediction in Spanish and English Tweets Using Monolingual and Multilingual BERT and Ensemble Models}
%
%
\author{Angel Felipe Magnossão de Paula\inst{1}\orcidID{0000-0001-8575-5012} \and
Roberto Fray da Silva\inst{2}\orcidID{0000-0002-9792-0553} \and
Ipek Baris Schlicht\inst{1}\orcidID{0000-0002-5037-2203}}
\authorrunning{A.F.M. Paula et al.}
%
\institute{Universitat Politècnica de València, Spain \\
\email{\{adepau, ibarsch\}@doctor.upv.es} \\
\and
Escola Politécnica da Universidade de São Paulo \\
\email{roberto.fray.silva@gmail.com}\\
}
\maketitle              
\let\thefootnote\relax\footnotetext{\textit{IberLEF 2021, September 2021, Málaga, Spain.}\\Copyright \textcopyright\ 2021 for this paper by its authors. Use permitted under Creative Commons License Attribution 4.0 International (CC BY 4.0).}
\begin{abstract}

The popularity of social media has created problems such as hate speech and sexism. The identification and classification of sexism in social media are very relevant tasks, as they would allow building a healthier social environment. Nevertheless, these tasks are considerably challenging. This work proposes a system to use multilingual and monolingual BERT and data points translation and ensemble strategies for sexism identification and classification in English and Spanish. It was conducted in the context of the sEXism Identification in Social neTworks shared 2021 (EXIST 2021) task, proposed by the Iberian Languages Evaluation Forum (IberLEF). The proposed system and its main components are described, and an in-depth hyperparameters analysis is conducted. The main results observed were: (i) the system obtained better results than the baseline model (multilingual BERT); (ii) ensemble models obtained better results than monolingual models; and (iii) an ensemble model considering all individual models and the best standardized values obtained the best accuracies and F1-scores for both tasks. This work obtained first place in both tasks at EXIST, with the highest accuracies (0.780 for task 1 and 0.658 for task 2) and F1-scores (F1-binary of 0.780 for task 1 and F1-macro of 0.579 for task 2).

\end{abstract}
\section{Introduction}
The emergence of social networks and microblogs has created a new medium for people to express themselves, providing freedom of speech and the possibility for quickly spreading opinions, news, and information \cite{lopes2014impact,jang2021study}. This has impacted considerably on peoples' lives, by increasing access to all kinds of information. Nevertheless, a small part of the users employs those media for spreading hate messages, increasing the impacts of racism, sexism, and others types of prejudices and hate speech \cite{wani2021impact,founta2018large}.


One crucial problem faced by the different stakeholders related to social media platforms is detecting hate speech \cite{founta2018large,yin2021towards,chiril2020annotated,chetty2018hate}, both in general and issue-specific forms. Also, some types of hate speech tend to be more challenging to identify, as they present characteristics such as irony or sarcasm, among others \cite{yin2021towards,chiril2020annotated,pamungkas2020misogyny}. Sexism is a type of toxic language that could be used both as hate speech and, in a much more subtle way, as sarcasm. Sexism is related to all kinds of behaviors and content that aim to spread prejudice against women, reduce their importance in society, or behave aggressively or offensively \cite{rodriguez2020automatic,chiril2020annotated,pamungkas2020misogyny}. There are several forms of sexism, and identifying them in social media messages is a fundamental challenge among the various natural language processing (NLP) tasks \cite{founta2018large,yin2021towards,rodriguez2020automatic,chiril2020annotated,pamungkas2020misogyny,chetty2018hate,poletto2020resources}.

The detection of sexism can be broken into two main tasks: (i) sexism identification, which aims to identify if a message or post containing sexist contents (regardless of the type of sexism contained in it); and (ii) sexism classification, which aims to classify the type of sexism contained in a given sexist message or post \cite{chiril2020annotated,rodriguez2020automatic,poletto2020resources,frenda2019online}. Both are very relevant, and the second task is dependent on the first, as it needs posts that are confirmed as sexist as inputs for the different classification models. Additionally, the difficulty of using data-driven models may increase for languages that are more complex or that have fewer resources available, such as high-quality word embeddings, pre-trained language-specific models, task-specific lexicons, among others.

To advance the state-of-the-art knowledge in both sexism identification and classification on social media messages,  the Iberian Languages Evaluation Forum (IberLEF) proposed the sEXism Identification in Social neTworks shared 2021 (EXIST 2021) shared task. For the rest of this work, this challenge will be referred to as EXIST shared task. The main goal of the IberLEF forum is to promote scientific advances towards innovative solutions for detecting sexism on social media platforms \cite{EXIST2021}. For this reason, the 2021 shared task provided datasets in English and Spanish for both tasks labeled by experts, following state-of-the-art data collection and labeling procedures \cite{EXIST2021}. Those datasets are expected to become benchmarks for state-of-the-art research on sexism identification and classification on social media messages.

Therefore, a relevant gap in the literature is to develop data-driven models that better identify and classify sexist content on social media messages, considering the implementation in different languages. This would: (i) advance both the knowledge on the use of artificial intelligent models for data-driven sexism identification and detection; (ii) provide a better methodology for identifying and classifying sexist content, which is highly relevant for identifying unacceptable user behavior; and (iii) address the problem of generalizing the model throughout different languages. Related to this gap, it is vital to observe that identifying online sexism can be considerably challenging because posts may have several forms: they may sound hateful and offensive, or friendly and funny, misleading the current classifying models used for this task \cite{rodriguez2020automatic}.

State of the art systems for addressing those tasks for multiple languages uses the Bidirectional Encoder Representations from Transformers (BERT) multilingual model, an NLP model that uses transformers and is pre-trained on a comprehensive text corpora \cite{sohn2019mc,pavlopoulos2019convai,mozafari2020hate,mozafari2019bert,koufakou2020hurtbert}. This model is trained on datasets of multiple languages, but it is not language-specific. The pre-trained models are then fine-tuned on task-specific datasets on the target language.

The main goal of this work is to propose and evaluate a system to identify and classify sexist content in social media messages in multiple languages, using the EXIST 2021 shared task dataset \cite{EXIST2021} for implementation and evaluation. The official shared task metrics were used:  accuracy for task 1 (sexism detection) and F1-macro for task 2 (sexism classification). However, we also implemented other relevant metrics for NLP tasks to better evaluate the different models in relation to the state of the art baseline model, the multilingual BERT: precision and recall.

The three main research questions that are going to be addressed in this work are: (i) does the use of monolingual BERT models provides better results than the multilingual BERT model to identify and classify sexist content on social media messages in English and Spanish?; (ii) does the use of an ensemble strategy improves the results of the individual models?; and (iii) does the results differ between the English and Spanish languages? Besides answering those three questions, this work also conducts an in-depth analysis of the main hyperparameters for the implemented models for both languages.

The main contribution of this work is to propose and evaluate the sexism identification and classification system in multiple languages considering different components: monolingual BERT models, multilingual BERT, data points translation, and different ensemble strategies. We also explore the main hyperparameters of the implemented models in-depth, comparing the final models with the state-of-the-art BERT multilingual model. This work obtained first place in both sexism identification and classification tasks at the EXIST shared task \cite{EXIST2021}.

This work is organized in the following sections: section \ref{section_2} describes the main concepts and models used for sexism prediction in social media messages; section \ref{section_3} contains the main steps of the methodology used; section \ref{section_4} describes the proposed system for addressing both sexism identification and classification; section \ref{section_5} contains the main results of the system's implementation on the EXIST shared task dataset; section \ref{section_6} contains a discussion of relevant topics on the system's use, modification, and potential improvements; and section \ref{section_7} concludes this paper.

\section{Sexism identification and classification using artificial intelligence models}\label{section_2}
The works by \cite{pamungkas2020misogyny,rodriguez2020automatic,chiril2020annotated} explore in-depth the impacts of the different types of sexism on social media platforms, describing several important classes of sexism. As sexism is an important type of hate speech, we also refer the reader to the works by \cite{chetty2018hate,yin2021towards} for excellent reviews on identifying and classifying the different forms of hate speech. The main concepts observed in those works were considered in our approach.

This work addresses two very relevant tasks: (i) sexism identification in natural language texts; and (ii) classification of types of sexism in natural language texts. Some examples of works that addressed the first task are \cite{mozafari2020hate,lynn2019comparison,pamungkas2018automatic}. Examples that addressed the second task are \cite{founta2018large,koufakou2020hurtbert,sharifirad2019learning}.

It is essential to observe that the second task is considerably more complex because different languages can be used in the different classes (as well as the traditional problems related to social media messages: abbreviations, emojis, misspellings, memes, among others). 

Although there are a variety of different models and strategies used for sexism detection and classification, the most traditionally used models are: support vector machines (SVM), convolutional neural networks (CNN), long short-term neural networks (LSTM), and BiLSTM \cite{pamungkas2020misogyny,istaiteh2020racist,rodriguez2020automatic,frenda2019online,chiril2020annotated,yin2021towards}. In the last years, the BERT has been widely used \cite{pamungkas2020misogyny,istaiteh2020racist,rodriguez2020automatic,frenda2019online,chiril2020annotated,yin2021towards}. This model (and its variations) have presented the best results on those tasks, as observed in the works by \cite{koufakou2020hurtbert,sohn2019mc,mozafari2020hate,pavlopoulos2019convai}.


The NLP literature addresses several identification and classification tasks related to extracting and evaluating opinions from natural language texts. In general, those tasks are addressed using three main approaches \cite{Johnman2018,Sohangir2018,Nassirtoussi2014}: (i) lexical-based, in which specific dictionaries (lists of words with corresponding values on important dimensions for the task) are used to classify the input text; (ii) statistical learning or machine learning-based, in which machine learning and deep learning models are used, generally with word embeddings or bag of words models, to classify the text; and (iii) hybrid, in which both lexicons and machine learning models are used. 

However, it is essential to note that: (i) lexical-based systems are not able to learn (and could be improved by using a deep learning model, such as BERT); (ii) deep learning models, especially the multilingual BERT, are state-of-the-art on sexism identification and classification \cite{sohn2019mc,pavlopoulos2019convai,mozafari2020hate,mozafari2019bert,koufakou2020hurtbert}; (iii) lexicons tend to be language-specific, making it more challenging to apply the solutions for multiple languages; and (iv) few works use BERT with domain-specific lexicons for sexism identification and classification. One of those lexicons that are highly relevant in this context is the Hurtlex lexicon \cite{bassignana2018hurtlex}. This was used in the works by \cite{pamungkas2018automatic,koufakou2020hurtbert}, among several others.  

The BERT model was proposed by \cite{devlin2018bert} and can be described as a language learning model aimed at providing a general structure that can be further refined with fine-tuning on specific tasks and domains. Its main objective is to learn the main features and semantics of a language, based on semi-supervised learning on a vast text corpora (such as the BookCorpus and the Wikipedia database) \cite{devlin2018bert,rogers2020primer,acheampong2021transformer}. Its architecture and training workflow is composed of three main components: transformers (which is an advanced deep learning model), bidirectional training, and use of encoder representations \cite{devlin2018bert,rogers2020primer,acheampong2021transformer}. 

In this work, we use the multilingual BERT \cite{devlin2018bert}, the English version of the model \cite{devlin2018bert}, and the Spanish version of the model, called BETO \cite{CaneteCFP2020}. For an in-depth analysis of how the BERT model works, we refer the reader to \cite{rogers2020primer}. For an in-depth comparison with multilingual BERT with other models, as well as an in-depth description of how they work, we refer the reader to \cite{wu2019beto}.

However, very few works in the literature consider dealing with datasets with multiple languages. This work addressed this gap by proposing a system that contains multiple models and ensemble strategies.

This paper aims to fulfill the gap of evaluating monolingual and multilingual BERT models for identifying and classifying sexism in texts in multiple languages.

\section{Methodology}\label{section_3}
\begin{figure*}[h!]
    \centering
    \includegraphics[scale=0.65]{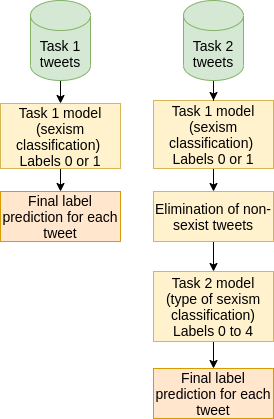}
    \caption{Illustration of the two main tasks evaluated in this research.}
    \label{fig:fig1}
\end{figure*}

The methodology used in this work was composed of six steps. Figure \ref{fig:fig1} illustrates the strategy used to tackle each of the tasks. For task 1 (sexism identification), the classification models considered two labels: 0 (non-sexist) and 1 (sexist). For task 2 (sexism classification), the tweets labeled as non-sexist (from task 1) were eliminated. Then, the classification models were used to predict the following sexism categories on the remaining tweets: ideological and inequality; stereotyping and dominance; objectification; sexual violence; and misogyny and non-sexual violence. For a thorough description of those classes, we refer to the EXIST shared task at IberLEF 2021 \cite{EXIST2021}, which developed and labeled the dataset that was used in this research.

The steps of the methodology were:


\textbf{1. Data collection}: we used the dataset developed for the EXIST shared task at IberLEF 2021 \cite{EXIST2021}. This dataset contained labeled data from two social media platforms: Twitter and Gab. For an in-depth description of this dataset, we refer the reader to Section \ref{section_5} of this work;

\textbf{2. Data processing}: for both tasks, we used the following processing techniques: separation of the dataset between languages (English and Spanish), tokenization, lemmatization, and elimination of stop words. These are widely used in the literature for the implementation of machine learning models on NLP tasks, such as hate speech detection, sexism identification, sentiment analysis, among others \cite{sohn2019mc,pavlopoulos2019convai,mozafari2020hate,mozafari2019bert,koufakou2020hurtbert,Nassirtoussi2014}. There was no need to eliminate data points from the datasets, as the shared task organizers had already thoroughly curated them. The training subset was then divided into training (80\%) and validation (20\%) for cross-validation purposes. Additionally, one of the training strategies used for some of the models implemented involved translating the social media messages from one language to the other (for example, from English to Spanish to train a Spanish language model). This strategy doubled the number of data points available for the single language models (even if part of the meaning may have been lost during the translation process). The googletrans (\url{https://github.com/ssut/py-googletrans}) library was used for the translation process;

\textbf{3. Exploratory data analysis}: in this step, an exploratory analysis of the dataset was conducted to understand better the different class distributions on both tasks throughout the training dataset. No data imbalance problems were observed;

\textbf{4. Model implementation and hyperparameters analysis}: in this research, we implemented the following models: (i) the BERT Multilingual model or mBERT \cite{devlin2018bert} (named M1 in this research); (ii) single language models (one for English and one for Spanish, named M2-English and M2-Spanish); (iii) single language models with translated data points (one for English and one for Spanish, named M3-English and M3-Spanish); and (iv) ensemble models (used only for the test subset). All the implementations were conducted with the Hugging Face BERT implementation library (https://huggingface.co/transformers/index.html) \cite{wolf2020transformers}, with 10-fold cross-validation on the training stage. A thorough hyperparameters analysis was conducted, considering the following hyperparameters and values: output BERT type (hidden or pooler), batch size (32 and 64), learning rate (0.00002, 0.00003, and 0.00005), and number of epochs (1 to 8). Following the official metrics of the EXIST 2021 shared task, accuracy was used as the quality metric for model training on task 1, and F1-macro was used on task 2. Besides this metric, an analysis of model overfitting was conducted for each model, based on charts that contained the models' accuracies on the different epochs;

\textbf{5. Final models implementation}: the final models and model ensembles were built using the best hyperparameters identified in Step 4. They were then trained on the whole training datasets (training plus validation subsets). Table \ref{tab:tab1} contains all the final models implemented: (i) M1: multilingual model; (ii) separated single language models without translation on the training datasets (M2, composed of M2-English and M2-Spanish) and with translation on the training datasets (M3, composed of M3-English and M3-Spanish); (iii) English single language model with translation only on the test subset (M4) and on training and test subsets (M5), both derived from the M3-English model; (iv) Spanish single language model with translation only on the test subset (M6) and on training and test subsets to Spanish (M7), both derived from the M3-Spanish model; (v) ensembles considering only the best models: E1 (majority vote), E2 (higher unstandardized value), and E3 (higher standardized value); and (vi) ensembles considering all the models: E4 (majority vote), E5 (higher unstandardized value), and E6 (higher standardized value);

\begin{table}[h]
\caption{Final models implemented and their characteristics}
\label{tab:tab1}
\begin{tabular}{|l|l|l|l|l|}
\hline
\multirow{2}{*}{Model} & \multirow{2}{*}{Multilingual} & \multicolumn{2}{l|}{Translation of data points} & \multirow{2}{*}{Observations}     \\ \cline{3-4}
                       &                               & Training               & Test                   &                                   \\ \hline
M1                     & \multicolumn{1}{c|}{X}        &                        &                        & Baseline model                    \\
M2                     &                               &                        &                        & Separated single language models  \\
M3                     &                               & \multicolumn{1}{c|}{X} &                        & Separated single language models  \\
M4                     &                               &                        & \multicolumn{1}{c|}{X} & English single language model     \\
M5                     &                               & \multicolumn{1}{c|}{X} & \multicolumn{1}{c|}{X} & English single language model     \\
M6                     &                               &                        & \multicolumn{1}{c|}{X} & Spanish single language model     \\
M7                     &                               & \multicolumn{1}{c|}{X} & \multicolumn{1}{c|}{X} & Spanish single language model     \\
E1                     & \multicolumn{3}{c|}{Depends on the individual models used}                      & Majority vote, best models        \\
E2                     & \multicolumn{3}{c|}{Depends on the individual models used}                      & Unstandardized value, best models \\
E3                     & \multicolumn{3}{c|}{Depends on the individual models used}                      & Standardized value, best models   \\
E4                     & \multicolumn{3}{c|}{Depends on the individual models used}                      & Majority vote, all models         \\
E5                     & \multicolumn{3}{c|}{Depends on the individual models used}                      & Unstandardized value, all models  \\
E6                     & \multicolumn{3}{c|}{Depends on the individual models used}                      & Standardized value, all models    \\ \hline
\end{tabular}
\end{table}

\textbf{6. Models comparison}: the final comparison of all models was then conducted on the test subsets. The official metrics for the EXIST shared task at IberLEF 2021 \cite{EXIST2021} were considered the quality metrics for both sexism identification and classification tasks. For task 1, we evaluated the accuracy, precision, recall, and F1-binary metrics. For task 2, we evaluated the accuracy, precision, recall, and F1-macro metrics). Additionally, an analysis of a sample of correctly and incorrectly classified data points on the test set was conducted, aiming to better understand each model's main strengths, weaknesses, and opportunities for future improvements. Lastly, the best model was chosen.

The implementation was done using Python on a Google Collaboratory Pro ({\url{https://colab.research.google.com/}}) TPU , with the following technical specifications: Intel(R) Xeon(R) CPU @ 2.30GHz CPU, 26GB of RAM, and TPU v2. The code implemented is available on an open Github repository ({\url{https://github.com/AngelFelipeMP/BERT-tweets-sexims-classification}}). Section \ref{section_5} presents the main results of the exploratory data analysis and the model implementations.

\section{Proposed system: components and implementation}\label{section_4}
\begin{figure*}[h!]
    \centering
    \includegraphics[width=\textwidth]{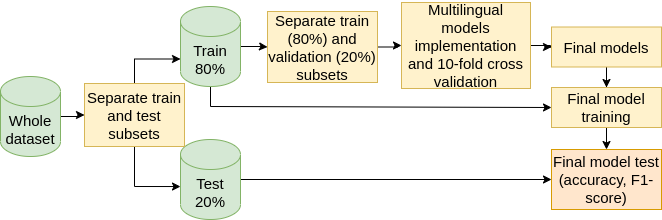}
    \caption{Workflow for training the Multilingual model (M1).}
    \label{fig:fig2}
\end{figure*}

\begin{figure*}[h!]
    \centering
    \includegraphics[width=\textwidth]{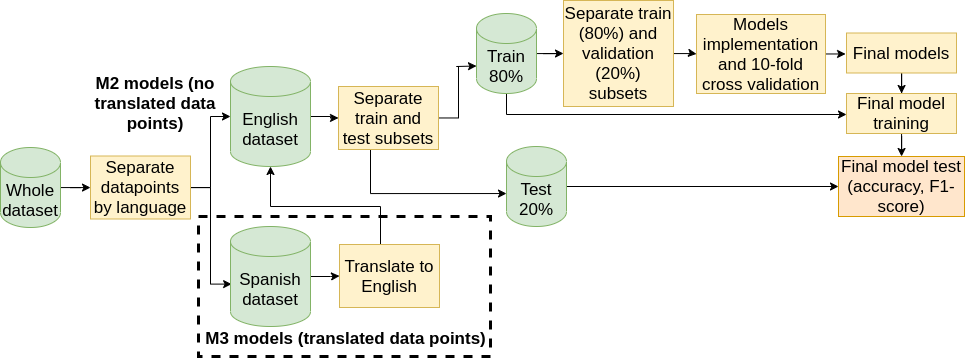}
    \caption{Workflow for training the monolingual models (M2-English and M2-Spanish) and the translated languages models (M3-English and M3-Spanish), considering the English models as an example.}
    \label{fig:fig3}
\end{figure*}

The proposed system considered two separate workflows: training and testing. The objective of the training workflow was to fine-tune the pre-trained BERT models. It considered three options: (i) using a multilingual BERT model (illustrated in Figure \ref{fig:fig2}), which was also considered our baseline, since it is the state of the art for multilingual NLP classification tasks; (ii) using monolingual BERT models without data points translation (illustrated in Figure \ref{fig:fig3}); and (iii) using monolingual BERT models with data points translation (illustrated in Figure \ref{fig:fig3}).

It is essential to observe that those three options considered 10-fold cross-validation on training to identify the best hyperparameter values for each model. The result from the first option was the M1 model. The results from the second option were the models M2-English and M2-Spanish, which would be used as components of the M2 model. The results from the third option were the models M3-English and M3-Spanish, which would be used as components of the M3 model.

\begin{figure*}[h!]
    \centering
    \includegraphics[scale=0.5]{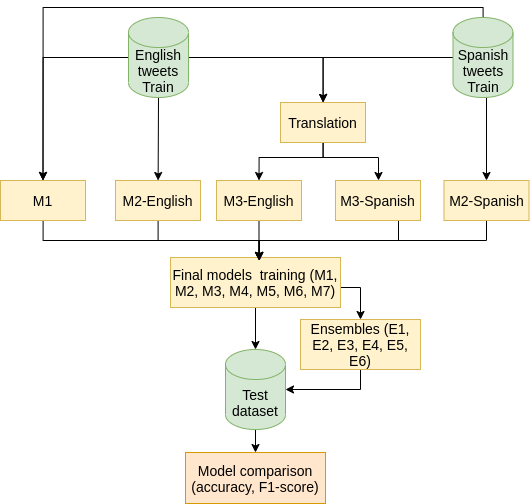}
    \caption{Workflow for testing the final models on the test subset.}
    \label{fig:fig4}
\end{figure*}

Figure \ref{fig:fig4} illustrates the test workflow. The objective of this workflow was to use the previously tested models as components for the final models, their training on the whole training dataset (training plus validation subsets), and testing on the test subset. This workflow also introduces the six ensemble models implemented, considering different model configurations and rules for generating the models. It is vital to observe that this model could be easily expanded for other languages, quality metrics, and data sources.

\section{Results}\label{section_5}


This section contains the main research results and is divided into three subsections: 5.1 contains a description of the dataset used; 5.2 contains the main results and observations related to the hyperparameters analysis; and 5.3 contains the final models' comparison on the test subset, considering four metrics: accuracy, precision, recall, and F1-score (F1-binary for task 1 and F1-macro for task 2).

\subsection{Description of the EXIST 2021 shared task dataset}
The dataset from EXIST 2021 shared task at IberLEF 2021 \cite{EXIST2021} was used in this work. This dataset contained labeled data from two social media platforms: (i) Twitter, with 6,977 tweets for training and 3,386 tweets for testing (both subsets equally distributed between English and Spanish); and (ii) Gab, with 492 gabs in English and 490 gabs in Spanish (used only for testing purposes). It is important to note that Gab is an uncensored social media website with considerably fewer users than Twitter.

It is vital to observe that the labeling procedure adopted by the shared task organizers considered both experts and crowdsourcing labeling (considering a specific procedure developed by experts in this domain). The dataset distribution was balanced on the training and test subsets. For a thorough description of the dataset, we refer the readers to IberLEF 2021 \cite{EXIST2021}.

The five classes that were used in this work for the sexism classification task (also referred to by the organizers of the dataset as sexism categorization) are the ones provided by the EXIST challenge dataset\cite{EXIST2021}. These classes contain, as described by \cite{EXIST2021,rodriguez2020automatic}:
\begin{itemize}
\item{Ideological and inequality: texts that affirm that the feminist movement deserves no credits, rejects the existence of inequality between genders, or claims that men are oppressed gender;}
\item{Stereotyping and dominance: texts that claim that women are inappropriate for specific tasks, suitable only for specific roles, or that men are superior to women;}
\item{Objectification: texts that claim that women should have certain physical qualities or that separate women from their dignity and personal aspects;}
\item{Sexual violence: texts that contain sexual suggestions or sexual harassment;}
\item{Misogyny and non-sexual violence: texts that express different forms of hatred and violence towards women.}
\end{itemize}

\subsection{Hyperparameters analysis}

Due to the considerable difference between the identification and classification tasks, this section will analyze both separately and then conclude with a comparison of the best hyperparameters values for all models for both tasks. The results observed in this subsection can be used as a guide for further implementations of BERT models for sexism identification and classification, considering multiple languages.

Table \ref{tab:tab2} illustrates the best hyperparameters values for task 1 for each of the five models implemented on the training step of the proposed system (M1, M2-Eng, M2-Sp, M3-Eng, and M3-Sp), as well as their associated quality metrics on the validation subset. 

\begin{table}[h]
\caption{Best hyperparameters values, accuracy, precision, recall and F1-binary for the different models on the validation subset of the training dataset for Task 1 - Sexism identification.}
\label{tab:tab2}
\begin{tabular}{|l|l|c|r|r|r|r|}
\hline
Lang.                    & Model           & \multicolumn{1}{l|}{Best hyperp. values}  & \multicolumn{1}{l|}{Acc.} & \multicolumn{1}{l|}{Prec.} & \multicolumn{1}{l|}{Rec.} & \multicolumn{1}{l|}{F1b} \\ \hline
Multi                    & M1              & OB:pooler / Lr:0.00005 / Bs:32 / Ne:7 & 0.774                     & 0.749                      & \textbf{0.808}            & \textbf{0.774}           \\ \hline
\multirow{2}{*}{English} & \textbf{M2-Eng} & OB:hidden / Lr:0.00005 / Bs:32 / Ne:5 & \textbf{0.782}            & \textbf{0.764}             & 0.788                     & 0.768                    \\
                         & M3-Eng          & OB:pooler / Lr:0.00005 / Bs:64 / Ne:6 & 0.765                     & 0.748                      & 0.778                     & 0.759                    \\ \hline
\multirow{2}{*}{Spanish} & \textbf{M2-Sp}  & OB:hidden / Lr:0.00005 / Bs:32 / Ne:8 & \textbf{0.790}            & \textbf{0.780}             & \textbf{0.795}            & \textbf{0.783}           \\
                         & M3-Sp           & OB:hidden / Lr:0.00005 / Bs:32 / Ne:6 & 0.775                     & 0.756                      & \textbf{0.795}            & 0.771                    \\ \hline
\end{tabular}
\hfill \break
Legend: Lang.: model language; OB: output BERT type; Lr: learning rate; Bs: batch size; Ne: number of epochs;  Acc.: accuracy; Prec.: precision; Rec.: recall; F1b: F1-binary.
\end{table}

Based on the analysis of Table \ref{tab:tab2}, it is important to observe that: (i) the M2-Eng (monolingual without translation) presented better results for the English language; (ii) the M2-Sp (monolingual without translation) presented better results for the Spanish language; (iii) the F1-binary for the Spanish language models (M2-Sp and M3-Sp) presented better results than for the English language models (M2-Eng and M3-Eng); (iv) most of the models presented better results by using the hidden output BERT type; (v) all models presented a learning rate of 0.00005; (vi) most models presented better results with a batch size of 32; and (vii) most models presented better results with 6 or more epochs. We have focused the analysis and model choice considering the accuracy, as it was the official metric for the EXIST shared task.

Table \ref{tab:tab3} illustrates the best hyperparameters values for task 2 for each of the five models implemented on the training step of the proposed system and their quality metrics on the validation subset. It is possible to observe that: (i) similar to task 1, the M1 model (multilingual) did not present better results than any of the languages; (ii) the M3-Eng (monolingual with translation) presented the best results for the English language; (iii) the M3-Sp (monolingual with translation) presented better results for the Spanish language; (iv) most of the models presented better results by using the hidden output BERT type; (v) like in task 1, all models presented better results by using a learning rate of 0.00005; (vi) most models presented better results with a batch size of 32; and (vii) most models presented better results with 7 or 8 epochs.

\begin{table}[]
\caption{Best hyperparameters values, accuracy, precision, recall and F1-macro for the different models on the validation subset of the training dataset for Task 2 - Sexism classification.}
\label{tab:tab3}
\begin{tabular}{|l|l|c|r|r|r|r|}
\hline
Lang.                    & Model           & \multicolumn{1}{l|}{Best hyperp. values}  & \multicolumn{1}{l|}{Acc.} & \multicolumn{1}{l|}{Prec.} & \multicolumn{1}{l|}{Rec.} & \multicolumn{1}{l|}{F1m} \\ \hline
Multi                    & M1-Multi        & OB:pooler / Lr:0.00005 / Bs:32 / Ne:8 & 0.636                     & 0.632                      & \textbf{0.624}            & \textbf{0.604}           \\ \hline
\multirow{2}{*}{English} & M2-Eng & OB:hidden / Lr:0.00005 / Bs:32 / Ne:8 & \textbf{0.661}            & \textbf{0.647}             & 0.633                     & 0.610                    \\
                         & \textbf{M3-Eng} & OB:hidden / Lr:0.00005 / Bs:32 / Ne:5 & \textbf{0.661}            & \textbf{0.660}             & \textbf{0.652}            & \textbf{0.632}           \\ \hline
\multirow{2}{*}{Spanish} & M2-Sp  & OB:hidden / Lr:0.00005 / Bs:32 / Ne:8 & \textbf{0.682}            & \textbf{0.656}             & \textbf{0.670}            & \textbf{0.628}           \\
                         & \textbf{M3-Sp}  & OB:hidden / Lr:0.00005 / Bs:64 / Ne:7 & 0.656                     & 0.653                      & \textbf{0.650}            & \textbf{0.630}           \\ \hline
\end{tabular}
\hfill \break
Legend: Lang.: model language; OB: output BERT type; Lr: learning rate; Bs: batch size; Ne: number of epochs;  Acc.: accuracy; Prec.: precision; Rec.: recall; F1m: F1-macro.
\end{table}

Lastly, Table \ref{tab:tab4} contains a cross-model and cross-language analysis of the results on the validation subset. It presents the hyperparameter values of the best models in each category (M1, M2-Eng, M2-Sp, M3-Eng, and M3-Sp), as a percentage of the total number of models, for each task. For example, on the first cell, it is possible to observe that, for the output BERT type on task 1, 60\% of the final models contained a hidden output BERT, while 40\% used the pooler type. Based on an analysis of this table, it is possible to conclude that: (i) for both tasks, the hidden output BERT type provided the best results; (ii) higher learning rates (0.00005) presented the best results for both tasks; (iii) the best batch size for both tasks was 32; and (iv) the best number of epochs differed among tasks, probably due to their different nature.

\begin{table}[]
\centering
\caption{Percentage of best performing models for each hyperparameter value for both tasks and languages, considering the highest accuracy for task 1 and F1-macro for task 2 on the validation subset.}
\label{tab:tab4}
\begin{tabular}{|l|l|l|}
\hline
\multirow{2}{*}{Hyperparameter} & \multicolumn{2}{l|}{\begin{tabular}[c]{@{}l@{}}Hyperparameter values and percentage \\ of best performing models\end{tabular}}                                                           \\ \cline{2-3} 
                                & Task 1                                                                                    & Task 2                                                                               \\ \hline
Output BERT type                & \textbf{\begin{tabular}[c]{@{}l@{}}Hidden: 60\%\\ Pooler: 40\%\end{tabular}}              & \textbf{\begin{tabular}[c]{@{}l@{}}Hidden: 80\%\\ Pooler: 20\%\end{tabular}}         \\ \hline
Learning rate                   & \begin{tabular}[c]{@{}l@{}}0.00002: 0\%\\ 0.00003: 0\%\\ 0.00005: 100\%\end{tabular}      & \begin{tabular}[c]{@{}l@{}}0.00002: 0\%\\ 0.00003: 0\%\\ 0.00005: 100\%\end{tabular} \\ \hline
Batch size                      & \textbf{\begin{tabular}[c]{@{}l@{}}32: 80\%\\ 64: 20\%\end{tabular}}                      & \textbf{\begin{tabular}[c]{@{}l@{}}32: 80\%\\ 64: 20\%\end{tabular}}                 \\ \hline
Number epochs                   & \textbf{\begin{tabular}[c]{@{}l@{}}\textless{}= 6: 60\%\\ 7: 20\%\\ 8: 20\%\end{tabular}} & \begin{tabular}[c]{@{}l@{}}\textless{}= 6: 20\%\\ 7: 20\%\\ 8: 60\%\end{tabular}     \\ \hline
\end{tabular}
\end{table}

\subsection{Final models comparison}

Table \ref{tab:tab5} contains the results of the thirteen final models on the test subset for both tasks, considering the following metrics: accuracy, precision, recall, and F1-score (F1-binary for task 1 and F1-macro for task 2). For both tasks, it is vital to observe that: (i) the ensemble models presented a better F1-score than the monolingual models and the multilingual model; and (ii) the E6 (ensemble model considering all individual models and the best standardized values) obtained the best accuracy and F1-score. 

\begin{table}[h!]
\centering
\caption{Accuracy, precision, recall and F1-score of the final models on the test subset for tasks 1 and 2.}
\label{tab:tab5}
\begin{tabular}{|l|c|c|c|l|l|l|l|l|}
\hline
\multirow{2}{*}{Model} & \multicolumn{4}{c|}{Task 1 - Sexism identification}                                                 & \multicolumn{4}{c|}{Task 2 - Sexism classification}               \\ \cline{2-9} 
                       & \multicolumn{1}{l|}{Acc.} & \multicolumn{1}{l|}{Prec.} & \multicolumn{1}{l|}{Rec.} & F1b            & Acc.           & Prec.          & Rec.           & F1m            \\ \hline
M1                     & 0.761                     & 0.739                      & 0.784                     & 0.761          & 0.621          & 0.617          & 0.621          & 0.611          \\
M2                     & 0.774                     & 0.749                      & \textbf{0.803}            & 0.775          & 0.688          & 0.677          & 0.674          & 0.675          \\
M3                     & 0.782                     & 0.773                      & 0.778                     & 0.775          & 0.676          & 0.661          & 0.663          & 0.658          \\
M4                     & 0.732                     & 0.737                      & 0.693                     & 0.715          & 0.612          & 0.607          & 0.597          & 0.595          \\
M5                     & 0.766                     & 0.773                      & 0.730                     & 0.751          & 0.661          & 0.648          & 0.657          & 0.649          \\
M6                     & 0.735                     & 0.739                      & 0.699                     & 0.718          & 0.639          & 0.629          & 0.621          & 0.624          \\
M7                     & 0.753                     & 0.734                      & 0.769                     & 0.751          & 0.642          & 0.625          & 0.627          & 0.624          \\
E1                     & 0.784                     & \textbf{0.797}             & 0.744                     & 0.769          & 0.669          & 0.683          & 0.666          & 0.653          \\
E2                     & 0.786                     & 0.777                      & 0.791                     & 0.781          & 0.686          & 0.673          & 0.677          & 0.674          \\
E3                     & 0.784                     & 0.773                      & 0.784                     & 0.779          & 0.682          & 0.669          & 0.674          & 0.670          \\
\textbf{E4}            & \textbf{0.790}            & 0.781                      & 0.787                     & \textbf{0.784} & 0.661          & 0.673          & 0.656          & 0.645          \\
E5                     & 0.785                     & 0.767                      & 0.799                     & 0.782          & 0.701          & 0.687          & 0.690          & 0.687          \\
\textbf{E6}            & \textbf{0.789}            & 0.776                      & 0.794                     & \textbf{0.785} & \textbf{0.703} & \textbf{0.690} & \textbf{0.692} & \textbf{0.689} \\ \hline
\end{tabular}
\hfill \break
\hfill \break
Legend:  Acc.: accuracy; Prec.: precision; Rec.: recall; F1b: F1b-binary; F1m: F1-macro. M1 is the baseline (BERT Multilingual).
\end{table}

For task 1, it is essential to observe on Table \ref{tab:tab5} that: (i) the baseline model (M1) presented better accuracy results than the M4, M6, and M7 models; (ii) the E4 model presented results that were comparable to the E6 model in terms of accuracy, both being considered the best models for this task; (iii) the E1 model presented the best precision; and (iv) the M2 model presented the best recall. For task 2, it can be observed that: (i) with an exception for M4, all models presented a better F1-macro than the M1 model, indicating that the use of monolingual may provide significantly better results than multilingual models for sexism classification; and (ii) the E6 model presented the best results for all metrics, indicating that it outperformed all other models for this task.

Table \ref{tab:tab6} presents a comparison of the best individual and ensemble models for tasks 1 and 2, considering the two official metrics of the EXIST shared task: accuracy and F1-score. Considering the differences of the F1-scores of each model and the best model, it is possible to conclude that: (i) the differences are significantly higher for the task of sexism classification; (ii) the baseline model (M1) obtained the worst F1-score among those models (around 3\% lower for task 1 and 11\% for task 2 in comparison to the E6 model); (iii) the baseline model (M1) obtained the worst accuracy for both tasks; and (iv) although the E4 model presented similar results for task 1, it observed a 6.39\% lower F1-score in comparison to the E6 model for task 2. The analysis of the models' accuracies lead to the same conclusions.

It is also possible to observe that the same overall conclusions hold for the accuracy metric.

\begin{table}[]
\caption{ Comparison of best individual and ensemble models for tasks 1 and 2 considering accuracy and F1-score.}
\label{tab:tab6}
\begin{tabular}{|l|c|c|l|l|l|l|l|l|}
\hline
\multirow{2}{*}{Model} & \multicolumn{4}{c|}{Task 1 - Sexism identification}                                 & \multicolumn{4}{c|}{Task 2 - Sexism classification} \\ \cline{2-9} 
                       & \multicolumn{1}{l|}{Acc.} & \multicolumn{1}{l|}{Diff E6} & F1b            & Diff E6 & Acc.             & Diff E6    & F1m              & Diff E6   \\ \hline
M1                     & 0.761                     & -3.55\%                      & 0.761          & -3.06\% & 0.621            & -11.66\%   & 0.611            & -11.32\%  \\
M2                     & 0.774                     & -1.90\%                      & 0.775          & -1.27\% & 0.688            & -2.13\%    & 0.675            & -2.03\%   \\
M3                     & 0.782                     & -0.89\%                      & 0.775          & -1.27\% & 0.676            & -3.84\%    & 0.658            & -4.50\%   \\
\textbf{E4}            & \textbf{0.790}            & 0.13\%                       & \textbf{0.784} & -0.13\% & 0.661            & -5.97\%    & 0.645            & -6.39\%   \\
\textbf{E6}            & \textbf{0.789}            & 0.00\%                       & \textbf{0.785} & 0.00\%  & \textbf{0.703}   & 0.00\%     & \textbf{0.689}   & 0.00\%    \\ \hline
\end{tabular}
\hfill \break
\hfill \break
Legend: Acc.: accuracy; Prec.: precision; Rec.: recall; F1b: F1b-binary; F1m: F1-macro. M1 is the baseline (BERT Multilingual). Diff E6 is the difference between that model’s metric and the same metric for the E6 model (best overall model for both tasks).
\end{table}

Our approach ranked first in both sexism identification and classification tasks at EXIST, with the highest accuracies (0.780 for task 1 and 0.658 for task 2) and F1-scores (F1-binary of 0.780 for task 1 and F1-macro of 0.579 for task 2), considering the E6 model. We also observed that ensemble models provide a better generalization.

\section{Discussions}\label{section_6}

This section briefly explores several important aspects related to the system proposed in this work and its results, encompassing the following topics: implementation aspects, system design, use of ensembles, system adaptation for other languages, results obtained concerning the literature, impacts of the different system components, and the use of the proposed system in real scenarios.

It is vital to note that the system proposed in this work can be extended using additional components with few adaptations to the code. Some additional interesting components to explore are lexicons (both generalist, such as Vader \cite{hutto2014vader} and domain-specific, such as Hurtlex \cite{bassignana2018hurtlex}), word embeddings, and transfer learning (via training on multiple datasets). Additional models could also be implemented to improve feature engineering (such as unsupervised learning models) or improve prediction quality (such as different weak models used in an ensemble strategy).

Although ensemble models are relatively common in other domains, such as price prediction \cite{Ballings2015} and sentiment analysis \cite{da2014tweet}, they are not widely spread on sexism identification and classification, as this is a new task. In general, if the behavior of the weak models can capture different aspects of the task, an ensemble strategy could improve the final prediction results \cite{Ballings2015,Dean2020}. In this work, we evaluated several simple average ensemble strategies. However, an in-depth analysis of more complex ensemble strategies with the proposed system could be conducted in future works. As was observed in this work, the use of ensembles can significantly improve the results obtained by the individual models. 

Another important aspect is related to adapting the proposed system for other languages. Concerning this aspect, it is vital to separate the languages into two main groups: (i) languages with individual BERT models already implemented; and (ii) languages that currently have no widely accepted individual BERT model implemented. In the first group, the system allows for easy implementation with minimal coding needed: the only components needed are the BERT individual model pre-trained on that language and the task-specific dataset for fine-tuning and testing. 

In the second group, it is necessary to train a language-specific BERT model before using the proposed system. This task demands a considerable amount of computation power and resources, demanding processing clusters and large text corpora in the target language (such as the Wikipedia text database). However, adaptations can be implemented in the proposed system to use different models that are easier to implement and demand fewer data, such as recurrent neural networks or convolutional neural networks with language-specific word embeddings or lexicons. The ensemble component can then be used, considering the BERT multilanguage (if it encompasses the target language) and the implemented models.

Lastly, the proposed system can be implemented and used in real case scenarios to improve sexism identification on social media platforms. After the hyperparameters and final models are chosen, as described and explored in this work, the prediction process is considerably fast. The system has the potential to be implemented as a separate service on the social media platform, analyzing the content published by its users and pointing out sexist messages and their respective classes (considering the five classes studied in this work).

\section{Conclusion and future work}\label{section_7}

As was explored throughout this work, a widespread problem on social networks and microblogs is the misuse of these tools to spread toxic language and sexist content. Identifying and detecting sexism in those media is considerably challenging, especially in a scenario with multiple languages. This paper explored the fine-tuning of multilingual and monolingual BERT models for English and Spanish and the use of different ensemble configurations to identify and classify sexism on tweets and gabs. The dataset used was provided by the EXIST shared task, which contained two tasks: sexism identification and sexism classification.

The proposed system in this research considered the use of fine-tuning of pre-trained BERT models and the translation of the training dataset (to increase the number of data points used by the model for learning) and ensemble models with different characteristics. Our central hypothesis was that using this system would provide better results than the traditional use of the multilingual BERT model. Our results have shown that the use of ensembles provided better results for both tasks, primarily the ensemble that considered all trained models and the higher standardized label values as the final predictions. This model obtained significantly better results than the baseline multilingual BERT model, with an F1-score around 3\% higher for the sexism identification task and 11\% higher for the sexism classification task. Those results and models and the in-depth hyperparameters analysis that was conducted can be used as a guide for future research on both tasks.

Future works are related to: (i) conducting an analysis considering additional datasets; (ii) implementing additional models; (iii) implementing different ensemble configurations; (iv) implementing unsupervised models for feature engineering; (v) analyzing the impacts on the models' results of using lexicons (both general and domain-specific) as features; (vi) analyzing the impacts on the models' results of using word embeddings as features; and (vii) implementing and evaluating the use of deep reinforcement learning to improve the models' results, especially on the sexism classification problem.

\bibliographystyle{splncs04}
\bibliography{reference}

\begin{thebibliography}{10}
\providecommand{\url}[1]{\texttt{#1}}
\providecommand{\urlprefix}{URL }
\providecommand{\doi}[1]{https://doi.org/#1}

\bibitem{acheampong2021transformer}
Acheampong, F.A., Nunoo-Mensah, H., Chen, W.: Transformer models for text-based
  emotion detection: a review of bert-based approaches. Artificial Intelligence
  Review pp. 1--41 (2021)

\bibitem{Ballings2015}
Ballings, M., Poel, D.V.D., Hespeels, N., Gryp, R.: {Evaluating multiple
  classifiers for stock price direction prediction}. Expert Systems with
  Applications  \textbf{42}(20),  7046--7056 (2015).
  \doi{10.1016/j.eswa.2015.05.013}

\bibitem{bassignana2018hurtlex}
Bassignana, E., Basile, V., Patti, V.: Hurtlex: A multilingual lexicon of words
  to hurt. In: 5th Italian Conference on Computational Linguistics, CLiC-it
  2018. vol.~2253, pp.~1--6. CEUR-WS (2018)

\bibitem{CaneteCFP2020}
Cañete, J., Chaperon, G., Fuentes, R., Ho, J.H., Kang, H., Pérez, J.: Spanish
  pre-trained bert model and evaluation data. In: PML4DC at ICLR 2020 (2020)

\bibitem{chetty2018hate}
Chetty, N., Alathur, S.: Hate speech review in the context of online social
  networks. Aggression and violent behavior  \textbf{40},  108--118 (2018)

\bibitem{chiril2020annotated}
Chiril, P., Moriceau, V., Benamara, F., Mari, A., Origgi, G., Coulomb-Gully,
  M.: An annotated corpus for sexism detection in french tweets. In:
  Proceedings of The 12th Language Resources and Evaluation Conference. pp.
  1397--1403 (2020)

\bibitem{da2014tweet}
Da~Silva, N.F., Hruschka, E.R., Hruschka~Jr, E.R.: Tweet sentiment analysis
  with classifier ensembles. Decision Support Systems  \textbf{66},  170--179
  (2014)

\bibitem{Dean2020}
Dean, N.E., y~Piontti, A.P., Madewell, Z.J., Cummings, D.A.T., Hitchings,
  M.D.T., Joshi, K., Kahn, R., Vespignani, A., Halloran, M.E., {Longini Jr},
  I.M.: {Ensemble forecast modeling for the design of COVID-19 vaccine efficacy
  trials}. Vaccine  \textbf{38}(46),  7213--7216 (2020)

\bibitem{devlin2018bert}
Devlin, J., Chang, M.W., Lee, K., Toutanova, K.: Bert: Pre-training of deep
  bidirectional transformers for language understanding. arXiv preprint
  arXiv:1810.04805  (2018)

\bibitem{founta2018large}
Founta, A., Djouvas, C., Chatzakou, D., Leontiadis, I., Blackburn, J.,
  Stringhini, G., Vakali, A., Sirivianos, M., Kourtellis, N.: Large scale
  crowdsourcing and characterization of twitter abusive behavior. In:
  Proceedings of the International AAAI Conference on Web and Social Media.
  vol.~12 (2018)

\bibitem{frenda2019online}
Frenda, S., Ghanem, B., Montes-y G{\'o}mez, M., Rosso, P.: Online hate speech
  against women: Automatic identification of misogyny and sexism on twitter.
  Journal of Intelligent \& Fuzzy Systems  \textbf{36}(5),  4743--4752 (2019)

\bibitem{hutto2014vader}
Hutto, C., Gilbert, E.: Vader: A parsimonious rule-based model for sentiment
  analysis of social media text. In: Proceedings of the International AAAI
  Conference on Web and Social Media. vol.~8 (2014)

\bibitem{istaiteh2020racist}
Istaiteh, O., Al-Omoush, R., Tedmori, S.: Racist and sexist hate speech
  detection: Literature review. In: 2020 International Conference on
  Intelligent Data Science Technologies and Applications (IDSTA). pp. 95--99.
  IEEE (2020)

\bibitem{jang2021study}
Jang, J.W., Park, Y.G., Hur, S.I., An, Y.J.: Study on the impact of
  activity-based flexible office characteristics on the employees' innovative
  behavioral intention. In: International Conference on Software Engineering,
  Artificial Intelligence, Networking and Parallel/Distributed Computing. pp.
  87--103. Springer (2021)

\bibitem{Johnman2018}
Johnman, M., Vanstone, B.J., Gepp, A.: {Predicting FTSE 100 returns and
  volatility using sentiment analysis}. Accounting \& Finance  \textbf{58},
  253--274 (2018). \doi{10.1111/acfi.12373}

\bibitem{koufakou2020hurtbert}
Koufakou, A., Pamungkas, E.W., Basile, V., Patti, V.: Hurtbert: Incorporating
  lexical features with bert for the detection of abusive language. In:
  Proceedings of the Fourth Workshop on Online Abuse and Harms. pp. 34--43
  (2020)

\bibitem{lopes2014impact}
Lopes, A.R.: The impact of social media on social movements: The new
  opportunity and mobilizing structure. Journal of Political Science Research
  \textbf{4}(1),  1--23 (2014)

\bibitem{lynn2019comparison}
Lynn, T., Endo, P.T., Rosati, P., Silva, I., Santos, G.L., Ging, D.: A
  comparison of machine learning approaches for detecting misogynistic speech
  in urban dictionary. In: 2019 International Conference on Cyber Situational
  Awareness, Data Analytics And Assessment (Cyber SA). pp.~1--8. IEEE (2019)

\bibitem{EXIST2021}
Montes, M., Rosso, P., Gonzalo, J., Aragón, E., Agerri, R., Álvarez Carmona,
  M.Ã., Mellado, E.Ã., Carrillo-de Albornoz, J., Chiruzzo, L., Freitas, L.,
  Adorno, H.G., Gutiérrez, Y., Zafra, S.M.J., Lima, S., Plaza-de Arco, F.M.,
  Taulé, M.e.: Proceedings of the iberian languages evaluation forum (iberlef
  2021). In: Proceedings of the Iberian Languages Evaluation Forum (IberLEF
  2021). CEUR Workshop Proceedings (2021)

\bibitem{mozafari2019bert}
Mozafari, M., Farahbakhsh, R., Crespi, N.: A bert-based transfer learning
  approach for hate speech detection in online social media. In: International
  Conference on Complex Networks and Their Applications. pp. 928--940. Springer
  (2019)

\bibitem{mozafari2020hate}
Mozafari, M., Farahbakhsh, R., Crespi, N.: Hate speech detection and racial
  bias mitigation in social media based on bert model. PloS one
  \textbf{15}(8),  e0237861 (2020)

\bibitem{Nassirtoussi2014}
Nassirtoussi, A.K., Aghabozorgi, S., Wah, T.Y., Ngo, D.C.L.: {Text mining for
  market prediction : a systematic review}. Expert Systems with Applications
  \textbf{41}(16),  7653--7670 (2014). \doi{10.1016/j.eswa.2014.06.009}

\bibitem{pamungkas2020misogyny}
Pamungkas, E.W., Basile, V., Patti, V.: Misogyny detection in twitter: a
  multilingual and cross-domain study. Information Processing \& Management
  \textbf{57}(6),  102360 (2020)

\bibitem{pamungkas2018automatic}
Pamungkas, E.W., Cignarella, A.T., Basile, V., Patti, V., et~al.: Automatic
  identification of misogyny in english and italian tweets at evalita 2018 with
  a multilingual hate lexicon. In: Sixth Evaluation Campaign of Natural
  Language Processing and Speech Tools for Italian (EVALITA 2018). vol.~2263,
  pp.~1--6. CEUR-WS (2018)

\bibitem{pavlopoulos2019convai}
Pavlopoulos, J., Thain, N., Dixon, L., Androutsopoulos, I.: Convai at
  semeval-2019 task 6: Offensive language identification and categorization
  with perspective and bert. In: Proceedings of the 13th international Workshop
  on Semantic Evaluation. pp. 571--576 (2019)

\bibitem{poletto2020resources}
Poletto, F., Basile, V., Sanguinetti, M., Bosco, C., Patti, V.: Resources and
  benchmark corpora for hate speech detection: a systematic review. Language
  Resources and Evaluation pp. 1--47 (2020)

\bibitem{rodriguez2020automatic}
Rodr{\'\i}guez-S{\'a}nchez, F., Carrillo-de Albornoz, J., Plaza, L.: Automatic
  classification of sexism in social networks: An empirical study on twitter
  data. IEEE Access  \textbf{8},  219563--219576 (2020)

\bibitem{rogers2020primer}
Rogers, A., Kovaleva, O., Rumshisky, A.: A primer in bertology: What we know
  about how bert works. Transactions of the Association for Computational
  Linguistics  \textbf{8},  842--866 (2020)

\bibitem{sharifirad2019learning}
Sharifirad, S., Jacovi, A., Univesity, I.B.I., Matwin, S.: Learning and
  understanding different categories of sexism using convolutional neural
  network's filters. In: Proceedings of the 2019 Workshop on Widening NLP.
  pp. 21--23 (2019)

\bibitem{Sohangir2018}
Sohangir, S., Wang, D., Pomeranets, A., Khoshgoftaar, T.M.: {Big data : deep
  Learning for financial sentiment analysis}. Journal of Big Data
  \textbf{5}(3),  1--25 (2018). \doi{10.1186/s40537-017-0111-6}

\bibitem{sohn2019mc}
Sohn, H., Lee, H.: Mc-bert4hate: Hate speech detection using multi-channel bert
  for different languages and translations. In: 2019 International Conference
  on Data Mining Workshops (ICDMW). pp. 551--559. IEEE (2019)

\bibitem{wani2021impact}
Wani, M.A., Agarwal, N., Bours, P.: Impact of unreliable content on social
  media users during covid-19 and stance detection system. Electronics
  \textbf{10}(1), ~5 (2021)

\bibitem{wolf2020transformers}
Wolf, T., Chaumond, J., Debut, L., Sanh, V., Delangue, C., Moi, A., Cistac, P.,
  Funtowicz, M., Davison, J., Shleifer, S., et~al.: Transformers:
  State-of-the-art natural language processing. In: Proceedings of the 2020
  Conference on Empirical Methods in Natural Language Processing: System
  Demonstrations. pp. 38--45 (2020)

\bibitem{wu2019beto}
Wu, S., Dredze, M.: Beto, bentz, becas: The surprising cross-lingual
  effectiveness of bert. arXiv preprint arXiv:1904.09077  (2019)

\bibitem{yin2021towards}
Yin, W., Zubiaga, A.: Towards generalisable hate speech detection: a review on
  obstacles and solutions. arXiv preprint arXiv:2102.08886  (2021)

\end{thebibliography}
\end{document}